\documentclass[letterpaper]{article} 
\usepackage{aaai22}  
\usepackage{times}  
\usepackage{helvet}  
\usepackage{courier}  
\usepackage[hyphens]{url}  
\usepackage{graphicx} 
\usepackage{natbib}  
\usepackage{caption} 
\usepackage{amsmath}
\usepackage{amssymb}
\DeclareCaptionStyle{ruled}{labelfont=normalfont,labelsep=colon,strut=off} 
\usepackage{algorithm}
\usepackage{algorithmic}
\usepackage{graphicx}
\usepackage{newfloat}
\usepackage{listings}
\usepackage{graphicx}
\usepackage{subfigure}
\usepackage{hyperref}

\hypersetup{hidelinks,
	colorlinks=true,
	allcolors=black,
	pdfstartview=Fit,
	breaklinks=true}

\floatstyle{ruled}
\newfloat{listing}{tb}{lst}{}
\floatname{listing}{Listing}
\title{Boosting Generative Zero-Shot Learning by Synthesizing Diverse Features with Attribute Augmentation}

\author {
	Xiaojie Zhao\textsuperscript{\rm 1}\thanks{Equal contribution.},
	Yuming Shen\textsuperscript{\rm 2}\footnotemark[1],
	Shidong Wang\textsuperscript{\rm 3},
	Haofeng Zhang\textsuperscript{\rm 1}\thanks{Corresponding author.}
}
\affiliations {
	\textsuperscript{\rm 1}School of Computer Science and Engineering, Nanjing University of Science and Technology, China\\
	\textsuperscript{\rm 2}Department of Engineering Science, University of Oxford, UK\\
	\textsuperscript{\rm 3}School of Engineering, Newcastle University, UK\\
	 zhaoxj@njust.edu.cn,  ym\_zmxncbv@hotmail.com,  shidong.wang@newcastle.ac.uk, zhanghf@njust.edu.cn 
}

\usepackage{bibentry}

\begin{document}
	
\maketitle
	
\begin{abstract}
The recent advance in deep generative models outlines a promising perspective in the realm of Zero-Shot Learning (ZSL).
Most generative ZSL methods use category semantic attributes plus a Gaussian noise to generate visual features. After generating unseen samples, this family of approaches effectively transforms the ZSL problem into a supervised classification scheme. However, the existing models use a single semantic attribute, which contains the complete attribute information of the category. The generated data also carry the complete attribute information, but in reality, visual samples usually have limited attributes. Therefore, the generated data from attribute could have incomplete semantics. Based on this fact, we propose a novel framework to boost ZSL by synthesizing diverse features. This method uses augmented semantic attributes to train the generative model, so as to simulate the real distribution of visual features. We evaluate the proposed model on four benchmark datasets, observing significant performance improvement against the state-of-the-art. Our codes are available
on \url{https://github.com/njzxj/SDFA}.

\end{abstract}

\section{Introduction}

In recent years, deep learning evolves rapidly. 
While it facilitates the utilization of vast data for modelling, a new problem appears that the training phase only covers limited scopes of samples, which requires an additional generalization stage to mitigate the gap between the seen concepts for training and the unseen ones during inference. Zero-Shot Learning (ZSL) then emerges accordingly. ZSL recognizes categories that do not belong to the training set through the auxiliary semantic attributes of sample categories \cite{zhang2019probabilistic}.

Existing ZSL approaches can be roughly divided into two categories according to training methods. One uses the mapping method to map between visual space and attribute space \cite{cacheux2019modeling,bucher2016improving,fu2015transductive,kodirov2017semantic,zhang2018triple}. The other method is the generation method \cite{wang2018zero,sun2020cooperative,narayan2020latent,yu2020episode}. It firstly trains the sample generator through the seen class instances and class attributes. Then, it caches the generated samples of the unseen class with the trained generator and the class attributes of the unseen classes. Finally, a held-out classifier can be trained together with the seen and synthesized unseen instances, mimicking a supervised classification scheme.

\begin{figure}[t]
	\centering
	\includegraphics[width=0.98\columnwidth]{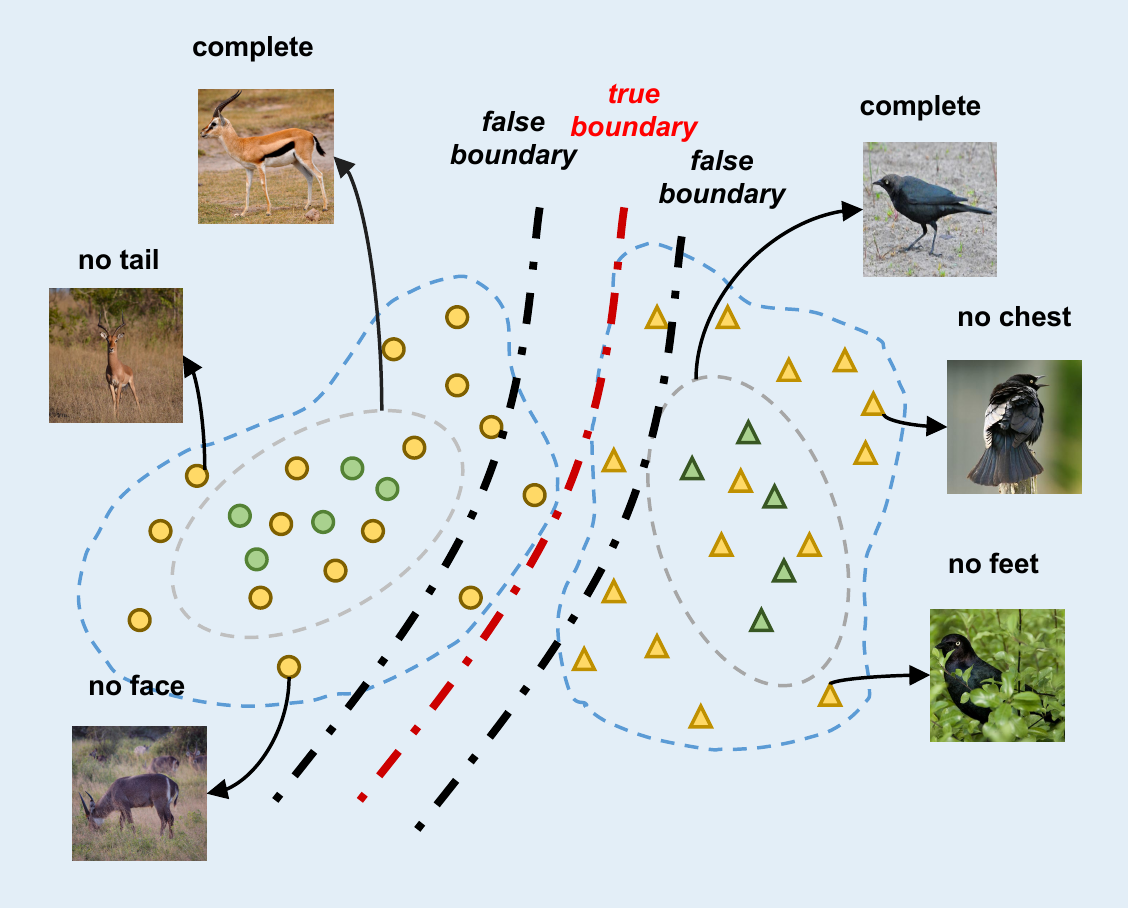} 
	\caption{Green represents the generated visual features, and yellow represents the real visual features. Traditional generative methods only use complete and single semantic attributes to generate visual features. This method cannot generate visual features with missing attributes. This makes the distribution of generated features and real features unable to fit. The classification boundary obtained from the visual features generated by single semantic attributes has great fluctuation. There is a big gap with the correct classification boundary.}
	\vspace{-3ex}
	\label{fig1}
\end{figure}

Our work mainly focuses on the generation method. When using semantic attributes to generate samples, the existing generation methods use complete semantic attributes, including all attribute information of corresponding class. Theoretically, when we generate visual features with complete semantic attributes, the generated visual features also contain complete attribute information, as depicted in Figure \ref{fig1}. However, the actual situation is that the visual features of categories do not always describe all the category attribute information. Let us imagine a picture of a horse in front. We cannot see the horse's tail, so the visual features extracted from this picture must lack the characteristic information of the tail. When generating pictures, the complete semantic attributes of horses are used, and the generated visual features must be complete, which must include the characteristics of tails. It is obviously contrary and harmful to classification learning as shown in Figure \ref{fig1}. The classification boundary obtained from the visual features generated by a single attribute has great fluctuation, which is different from the real classification boundary, will lead to the sample being misclassified.

In order to simulate the real situation and generate a variety of visual features, so that the generated visual features contain the class-level attribute information of various situations, we propose Synthesizing Diverse Features with Attribute Augmentation (SDFA$^{2}$). When generating visual features, our method better utilizes incomplete semantic attributes, which selectively masks out some bins of the semantic attributes. This simulates the absence of some attributes of visual features of real samples, so as to generate visual features closer to the real situation. Taking the semantic attribute of horse as an example, the dimension representing the tail in the semantic attribute of horse is set to 0 to represent the lack of tail attribute. The visual features of the horse generated by this incomplete semantics will also lose the attribute information of the tail. By means of this, the visual features generated by unseen classes can be closer to the real ones, and the held-out classifier trained on them observes realistic patterns, leading to better ZSL performance.

Meanwhile, the missing attribute information of visual features should be intelligible. As per our horse case, the missing tail of horse should be recognized. We add a self-monitoring module to set a Self-supervision label for each sample generated by incomplete semantic attributes to mark the missing attribute information. The generated samples are classified according to this label.

Our contributions can be described as follows:

\begin{itemize}
\item On the basis of using complete semantic attributes to generate visual features, we propose a novel framework SDFA$^{2}$ as a unified add-on to boost generative ZSL models. This method simulates the distribution of visual features in the real situation, which increases the authenticity of the generated features and improves the accuracy.

\item We additionally introduce self-supervised learning into our model, so that the generated visual features can be classified on the missing attributes.

\item We propose a general diversity feature generation method. We validate our method on two classical and two latest generative models, and the corresponding performance is improved significantly.
\end{itemize}

\section{Related Work}

\subsection{Zero-Shot Learning}
ZSL \cite{lampert2009learning,palatucci2009zero} aims to transfer the model trained by the seen classes to the unseen ones, usually with a semantic space between visible classes and invisible classes. ZSL is divided into Conventional ZSL (CZSL) and generalized ZSL (GZSL). CZSL \cite{zhang2015zero,fu2015zero,xian2016latent,frome2013devise} only contains unseen classes in the test phase while GZSL \cite{chao2016empirical,xian2018zero} contains seen classes and unseen classes. GZSL has attracted more attention because it describes a more realistic scenario. In the early ZSL, the mapping method is generally used to find the relationship between visual space and semantic space. Some map visual space to semantic space \cite{palatucci2009zero,romera2015embarrassingly}. Some project visual features and semantic features into a shared space \cite{liu2018generalized,jiang2018learning}. The others consider the mapping from the semantic space to the visual one to find classification prototypes \cite{2018Preserving,2017Learning}.

With the emergence of generation methods, semantic based visual feature generation Zero-Shot method also comes into being. F-CLSWGAN \cite{xian2018feature} generates unseen visual features by generative adversarial networks. F-VAEGAN-2D \cite{xian2019f} combines generative adversarial networks and VAE \cite{kingma2013auto}. LisGAN \cite{li2019leveraging} can directly generate the unseen features from random noises which are conditioned by the semantic descriptions. Cycle-CLSWGAN \cite{felix2018multi} proposes cycle consistency loss for cycle consistency detection. CE-GZSL \cite{han2021contrastive} adds contrastive learning for better instance-wise supervision. RFF-GZSL \cite{han2020learning} extracts the redundant features of the picture. ZSML \cite{verma2020meta} adds Meta-Learning to the training process. IZF \cite{shen2020invertible} learns factorized data embeddings with the forward pass of an invertible flow network, while the reverse pass generates data samples.

\begin{figure*}[ht]
	\centering
	\includegraphics[width=0.98\textwidth]{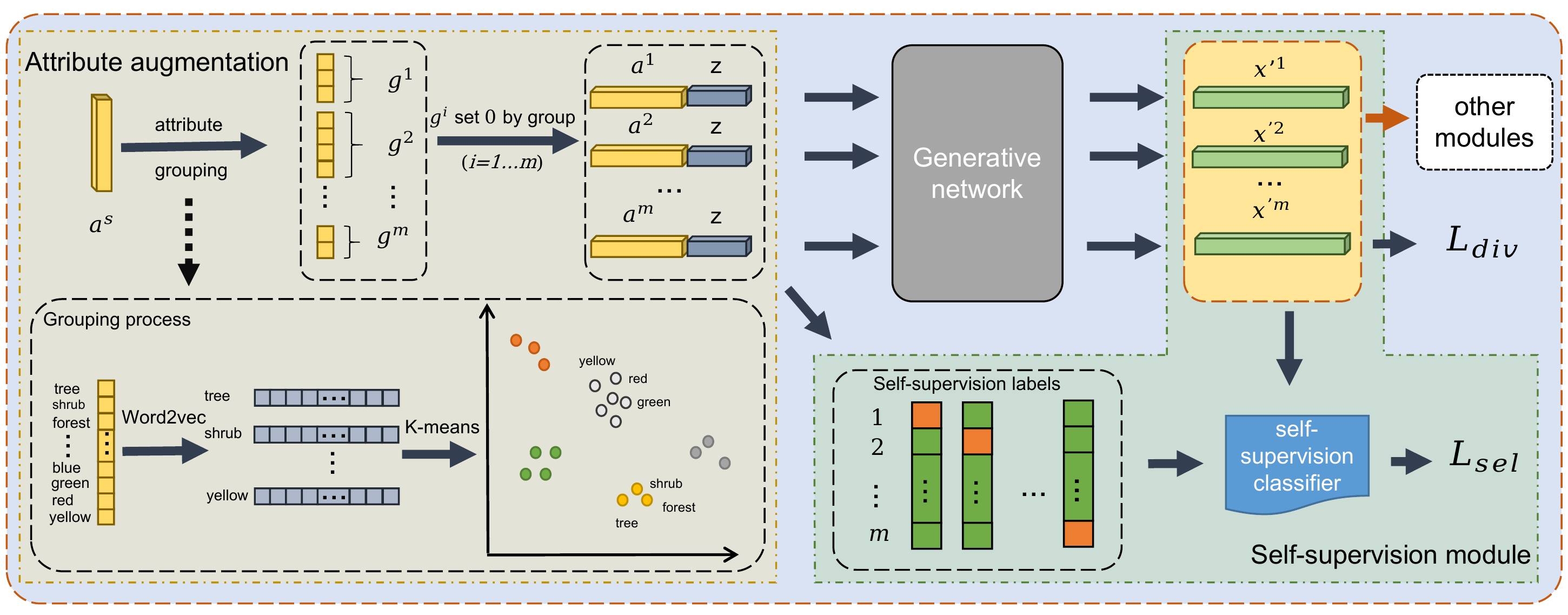} 
	\vspace{-2ex}
	\caption{The schematic of SDFA$^{2}$. We first group attributes, and then make attribute augmentation through setting 0 by group. The obtained attributes are used to generate diversity visual features. The generator and discriminator in generative network are replaceable for different generation networks. Other modules are replaceable network modules that downstream tasks of the generate network. For different GZSL models, other modules are different. At the same time, self-supervision labels are added to the generated visual features for self-supervision classification.}
	\label{fig2}
	\vspace{-3ex}
\end{figure*}

\subsection{Single Attribute Problem}
In the conventional ZSL setting, each category has a unified attribute, and there is a many-to-one relationship between the visual and semantic space. However, different visual features of the same category should also vary in semantic space. Based on this intuition, previous research has dealt with semantics in non generative methods. LDA \cite{li2018discriminative} claims that visual space may not be fully mapped to an artificially defined attribute space. Some use adaptive graph reconstruction scheme to excavate late semantics \cite{ding2019marginalized}. In order to avoid equal treatment of attributes, LFGAA \cite{liu2019attribute} proposes a practical framework to perform object-based attribute attention for semantic disambiguation. SP-AEN \cite{chen2018zero} solves the problem of semantic attribute loss. 

Most of the existing generation methods directly use the semantic attributes of the category when generating samples, and the generated visual features contain all the attribute information of the category. The reality is that some attribute features mismatch the corresponding images, and the visual features extracted by this method also have missing attributes. If we do not respect this objective fact and use the complete semantics of attribute information to generate unseen visual features with complete attribute information, the trained classification model must not be sensitive enough to those test samples with missing attribute features. Based on this fact, this paper proposes a Zero-Shot Learning method for incomplete semantic generation.

\subsection{Self-Supervised Learning}
Self-Supervised Learning (SSL) learns useful feature representations as a pretext to benefit the downstream tasks. Existing models deliver this with different approaches. Some methods process image color \cite{larsson2017colorization}. Some disrupt the arrangement order after cutting pictures \cite{santa2017deeppermnet}, while the others rotate images \cite{gidaris2018unsupervised,li2020international}. SSL can learn valuable representations. The purpose of our method is to use diversity attributes to generate missing features. In other words, the generated features hope to be distinguishable. Naturally, we integrate SSL into the network.

\section{Methodology}

\subsection{Problem Definition}
Suppose we have $S$ seen classes for training, and $U$ unseen classes that are only used for test. Let's use $Y_{s}$ and $Y_{u}$ to represent seen classes space and unseen classes space, $Y_{s}\bigcap Y_{u}=\varnothing $. $X_{S}=\left \{ x_{1},x_{2},\dots ,x_{S} \right \}\subset \mathbb{R}^{d_{x}\times S}$ is the visual feature of the sample in seen class, $Y_{S}=\left \{ y_{1},y_{2},\dots,y_{S} \right \}\subset \mathbb{R}^{1\times S}$ is the corresponding sample category label. Similarly, we let $X_{u}=\left \{ x_{S+1},x_{S+2},\dots,x_{S+U} \right \}\subset  \mathbb{R}^{d_{x}\times U}$ denotes the visual feature of the sample in unseen class, $Y_{U}=\left \{ y_{S+1},y_{S+2},\dots,y_{S+U} \right \}\subset \mathbb{R}^{1\times U}$ is the corresponding sample category label. At the same time, the semantic attribute $A=\left \{ a_{1},a_{2},\dots,a_{S},a_{S+1},\dots ,a_{S+U} \right \}\subset \mathbb{R}b^{d_{a}\times S+U}$ of all classes is available, where $A_{s}=\left \{ a_{1},a_{2},\dots ,a_{S} \right \}$ denotes the $S$ seen classes attribute and $A_{u}=\left \{ a_{S+1},a_{S+2},\dots ,a_{S+U} \right \}$ is the $U$ unseen classes. In GZSL, we purpose to get a classifier $f:X\rightarrow Y_{s}\cup Y_{u}$, to make $X$ which is from both seen classes and unseen classes is correctly classified.

\subsection{Attribute Augmentation}
When generating visual features, the traditional method uses the single semantic attributes of the class, which contains all the attribute information of the class. The generation of visual features with the complete semantic attributes of the class will theoretically include all the attribute information of the class. However, the actual situation is that the visual features of some instances may not include all attribute features of this category. For example, the visual features extracted from a front picture of a antelope must not contain the attribute information of the tail, and the visual features extracted by a bird standing on a tree with its feet covered by leaves must not contain the attribute information of the talons. Then the semantic attribute corresponding to its visual features must also lack the information of tail for antelope or talons for bird.

If we use single semantic attributes when generating visual semantic features, we will not be able to simulate the visual features extracted from pictures with incomplete attributes. If all visual features are generated by using single semantic attributes, the distribution of real visual features cannot be generated. A fact is that there must be a certain number of pictures lacking complete semantic information in the sample, and there must be visual features of the sample with missing attributes. When generating unseen class visual features, if the generated visual features are complete and the test instances of unseen classes contain incomplete information, the classification results will inevitably have deviation. The classification boundary obtained by using single semantic attributes to generate visual features has a large fluctuation range and is not accurate enough. In other words, if the missing attribute information is not considered, the trained classifier will not be sensitive to the missing attribute samples. The model is not robust enough, so as to reduce the accuracy.

In order to simulate this real situation, we generate visual features with complete attribute features and missing information at the same time. For class-level semantic attribute $a$, we artificially set dimensions to 0 to indicate the lack of attribute information. For example, the dimension of the tail in the semantic attribute of a horse is set to 0 to represent the lack of tail attribute information and the dimension of the feet in the semantic attribute of a bird is set to 0 to represent the lack of feet attribute information.

Because the semantic attribute is not marked for a specific class, but marked in the semantic space of all class, some dimensions of the semantic attribute itself are 0 for a class. Take ``antelope" as an example, the dimension representing ``ocean" in its semantic attribute itself is 0. Setting dimension ``ocean" to 0 is meaningless and cannot represent the incomplete semantic attribute of category ``antelope" without dimension ``ocean". The ``grassland" attribute of antelope is not 0. ``Grassland" and ``ocean" are both living environments, which are semantically similar. Another example is ``bipedal" and ``quadrapedal" are two dimensions descriptions of semantic attributes. In fact, they are different descriptions of the same part. We know, one class cannot have both ``bipedal" and ``quadrapedal" attributes.Another reason is that some datasets have higher dimensions of attributes. If each dimension is set to 0 in turn, the time and space overhead will be large.

So it is harmful to set each dimension of the attribute to 0 indiscriminately. With those in mind, we believe that attributes should be grouped according to similarity.  We divide the attributes by category. For each dimension description of attribute, we obtain the corresponding word vector $S=\left\lbrace s^{1},s^{2},\dots,s^{j}\right\rbrace$ through Word2vec \cite{2013Efficient} where $j$ is the number of attribute dimensions. Determine the number of clusters $m$. Our goal is to obtain the cluster $\left\lbrace g^{1},g^{2},\dots,g^{m}\right\rbrace $ by using expectation–maximization (EM) algorithm \cite{anzai2012pattern} to minimize the square error $E$:
\begin{align}
E= \sum_{i=1}^{m}\sum_{s\in g^{i}}\left \| s-\mu _{i} \right \|,
\end{align}
where $\mu _{i}$ is the mean vector of cluster $g^{i}$, and the calculation formula is:
\begin{align}
\mu _{i}=\frac{1}{\left | g^{i} \right |}\sum _{s\in g^{i}}s.
\end{align}

When getting attributes divided into $m$ groups, we denote $a^{i}$, with $i=1\dots m$, as the incomplete attributes which set the i-th group attribute of semantic attribute $a$ to 0. The visual features generated in this way which uses incomplete attributes are equivalent to simulating the front picture of a horse and the picture of a bird whose feet are covered by leaves. The diversity loss can be expressed as :

\begin{align}
L_{div-gan}=\frac{1}{m}\sum_{i=1}^{m} \left[ \varphi \left ( {x^{i}}' \right ) +\psi \left ({a^{i}} \right ) \right],
\end{align}
where ${x^i}'$ is the visual feature generated by $a^i$ and $\varphi \left ( \cdot  \right )$ represents the loss function related to visual features and $\psi \left ( \cdot  \right )$ represents the loss function related to semantic attributes in generative network.
Take WGAN as an example, the diversity losses can be expressed as: 
\begin{align}
L_{div-gan}=\mathbb{E}\left [ D\left ( x,a \right ) \right ]-\frac{1}{m}\sum_{i=1}^{m}\left ( \mathbb{E}\left [ D\left ( {x^i}',{a^{i}} \right )\right ]
\notag\right.
\\
\phantom{=\;\;}
\left. -\lambda \mathbb{E}\left [ \left ( \left \| \nabla_{\check{x^{i}}'}D\left ( \check{x^{i}}',a \right ) \right \|-1 \right )^{2} \right ] \right ),
\end{align}
where $a$ is the semantic attribute with $x$, ${x^i}'=G\left ( a^i,z \right )$ is the generated visual feature with $z\sim N\left ( 0,1 \right )$, $a^i$ is obtained from $a$. The last one is the gradient penalty, where$ \check{x}=\alpha x+\left ( 1-\alpha  \right ){x^i}'$ with $\alpha \sim U\left ( 0,1 \right )$ and $\lambda $ is the penalty coefficient.

\begin{table*}[t]
	\centering
	\resizebox{\textwidth}{!}{
	\begin{tabular}{l|lll|lll|lll|lll}
		\multicolumn{1}{c|}{} & \multicolumn{3}{c|}{\textbf{AWA}}                                                                                            & \multicolumn{3}{c}{\textbf{CUB}}                                                                                             & \multicolumn{3}{c|}{\textbf{aPY}}                                                                                            & \multicolumn{3}{c}{\textbf{SUN}}                                                                                            \\
		\multicolumn{1}{c|}{} & \multicolumn{1}{c}{\textit{\textbf{U}}} & \multicolumn{1}{c}{\textit{\textbf{S}}} & \multicolumn{1}{c|}{\textit{\textbf{H}}} & \multicolumn{1}{c}{\textit{\textbf{U}}} & \multicolumn{1}{c}{\textit{\textbf{S}}} & \multicolumn{1}{c|}{\textit{\textbf{H}}} & \multicolumn{1}{c}{\textit{\textbf{U}}} & \multicolumn{1}{c}{\textit{\textbf{S}}} & \multicolumn{1}{c|}{\textit{\textbf{H}}} & \multicolumn{1}{c}{\textit{\textbf{U}}} & \multicolumn{1}{c}{\textit{\textbf{S}}} & \multicolumn{1}{c}{\textit{\textbf{H}}} \\ \hline
		f-CLSWGAN \cite{xian2018feature}           & 57.9                                    & 61.4                                    & 59.6                                     & 43.7                                    & \textbf{57.7}                                    & 49.7                                     & 32.9                                    & 61.7                                    & 42.9                                     & 42.6                                    & 36.6                                    & 39.4                                    \\
		f-CLSWGAN+SDFA$^{2}$        & \textbf{59.1}                                    & \textbf{72.8}                                   & \textbf{65.2}                                    & \textbf{51.5}                                    & 57.5                                    & \textbf{54.3}                                     & \textbf{38.0}                                    & \textbf{62.8}                                    & \textbf{47.4}                                     & \textbf{48.7}                                   & \textbf{36.9}                                    & \textbf{42.0}                                    \\ \hline
		f-VAEGAN-D2 \cite{xian2019f}           & \textbf{59.7}                                    & 68.1                                    & 63.7                                     & 51.7                                    & \textbf{55.3}                                    & 53.5                                     & 29.1                                    & \textbf{60.4}                                    & 39.2                                     & \textbf{48.6}                                    & 36.7                                    & 41.8                                    \\
		f-VAEGAN-D2+SDFA$^{2}$      & 59.5                                      & \textbf{71.4}                                      & \textbf{64.7}                                       & \textbf{53.3}                                      & 55.0                                      & \textbf{54.2}                                       & \textbf{39.4}                                     & 57.5                                     & \textbf{46.8}                                      & 48.5                                      & \textbf{37.6}                                      & \textbf{42.4}                                      \\ \hline
		RFF-GZSL \cite{han2020learning}              & 49.7                                      & 50.8                                      & 61.5                                       & 52.1                                    & 57.0                                    & 54.4                                     & 18.6                                    & \textbf{86.7}                                    & 30.6                                     & 42.7                                    & 37.4                                    & 39.9                                    \\
		RFF-GZSL+SDFA$^{2}$         & \textbf{57.0}                                    & \textbf{75.5}                                    & \textbf{65.0}                                     & \textbf{52.3}                                      & \textbf{57.4}                                      & \textbf{54.7}                                       & \textbf{27.4}                                    & 66.9                                    & \textbf{38.9}                                     & \textbf{42.9}                                    & \textbf{51.6}                                    & \textbf{40.1}                                    \\ \hline
		CE-GZSL \cite{han2021contrastive}               & 57.0                                      & 74.9                                      & 64.7                                       & 49.0                                      & 57.4                                      & 53.0                                       & 9.58                                     & \textbf{88.4}                                     & 17.3                                      & 40.9                                      & \textbf{35.4}                                      & 37.9                                      \\
		CE-GZSL+SDFA$^{2}$          & \textbf{59.3}                                    & \textbf{75.0}                                    & \textbf{66.2}                                     & \textbf{59.2}                                    & \textbf{59.6}                                    & \textbf{54.0}                                     & \textbf{21.5}                                     & 85.1                                     & \textbf{34.3}                                      & \textbf{46.2}                                      & 32.6                                      & \textbf{38.2}                                      \\ \hline
	\end{tabular}
	}
	\vspace{-2ex}
	\caption{The table shows the results of SDFA$^{2}$ acting on four generative GZSL. Where $U$ represents the Top-1 accuracies of the unseen class, $S$ represents the Top-1 accuracies of the seen class, and $H$ is the harmonic average of the two. Each method is divided into two parts, in which the upper one is the result obtained from the original model and the lower one is the result obtained from the model after using SDFA$^{2}$.}
	\vspace{-3ex}
	\label{table1}
\end{table*}

\subsection{Self-Supervision of Missing Attributes}
The visual features of real instances can distinguish the missing semantic attributes, that is, the visual features extracted from the front picture of the horse can distinguish the missing tail attribute. Similarly, the visual features extracted from a picture of a bird whose feet are covered by leaves can identify the absence of feet attributes.

In order to make the visual features generated by diversity attributes more in line with the real situation, in other words, we can judge the missing attribute information of the generated visual features, we introduce self-supervision loss, As shown in Figure \ref{fig2}.
For the incomplete semantic attribute $a^{i}$ whose i-th group attribute is set to 0, we set a Self-supervision  label $h_{i}$ for the generated ${x^i}'=G\left ( a^i,z \right)$, to denote the visual features generated missing the attributes of the i-th group. Note that ${x^0}'$ represents visual features generated using complete semantic attributes $h_{0}$.

For each incompletely generated sample, we train a mapping relation $f_{\theta_{h}}:x\rightarrow h_{i}$, where $f_{\theta_{h}}$ is a learnable incomplete attribute recognition classifier with $\theta_{h}$. So the diversity identification loss is:
\begin{equation}
L_{self}=-\mathbb{E}\left [ logP\left ( h_{i}|{x^i}';\theta  \right ) \right ].
\end{equation}

\subsection{Total Loss}
Finally, according to the Eq.(3) and Eq.(5), diverse feature synthesis method consists of the loss of diversity semantic generation and the loss of missing attribute self-supervision. The total loss of our model is:
\begin{equation}
L_{div}=\lambda_{d}L_{div-gan}+\lambda_{s}L_{self},
\end{equation}
where $\lambda_{d}$ and $\lambda_{s}$ are hyper-parameters.

Figure \ref{fig2} is the overall framework of our method. Among them, generator and discriminator of generative network can be replaced with different types. Other modules are additional contents added outside the main body of the generated  network, such as classification loss in f-CLSWGAN, contrast loss in CE-GZSL, etc. Our method is a general module, which can be added to any ZSL that uses attributes to generate visual features.

\subsection{Training Classifier}
Compared with traditional method which only uses a single semantic attribute when generating visual features, our method additional uses the semantic attribute of attribute dimension grouped and setted 0 when generating visual features. Specifically, compared with the number of visual feature generation using a single semantic attribute, we generate visual features for each diverse semantic attribute in a certain proportion. We use the visual feature generated by single semantic attribute and the visual feature generated by diversity semantic attribute, which denotes ${X^U}'$, to train the classifier together with the visual feature $X^S$.

\begin{figure*}[ht]
	\subfigure[AWA1]{
		\begin{minipage}[t]{0.25\textwidth}
			\centering
			\includegraphics[width=1\columnwidth]{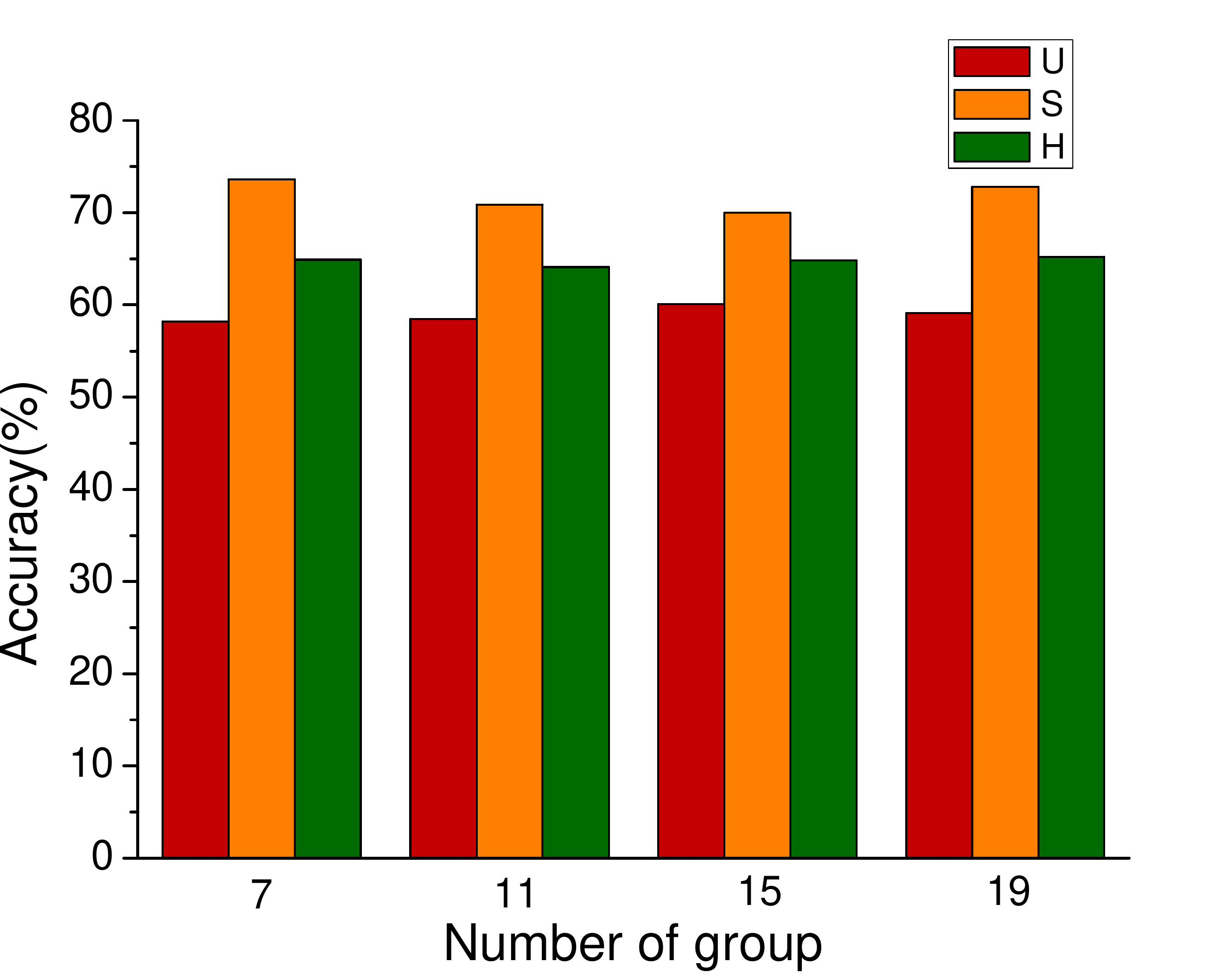}
		\end{minipage}%
	}%
	\subfigure[aPY]{
		\begin{minipage}[t]{0.25\textwidth}
			\centering
			\includegraphics[width=1\columnwidth]{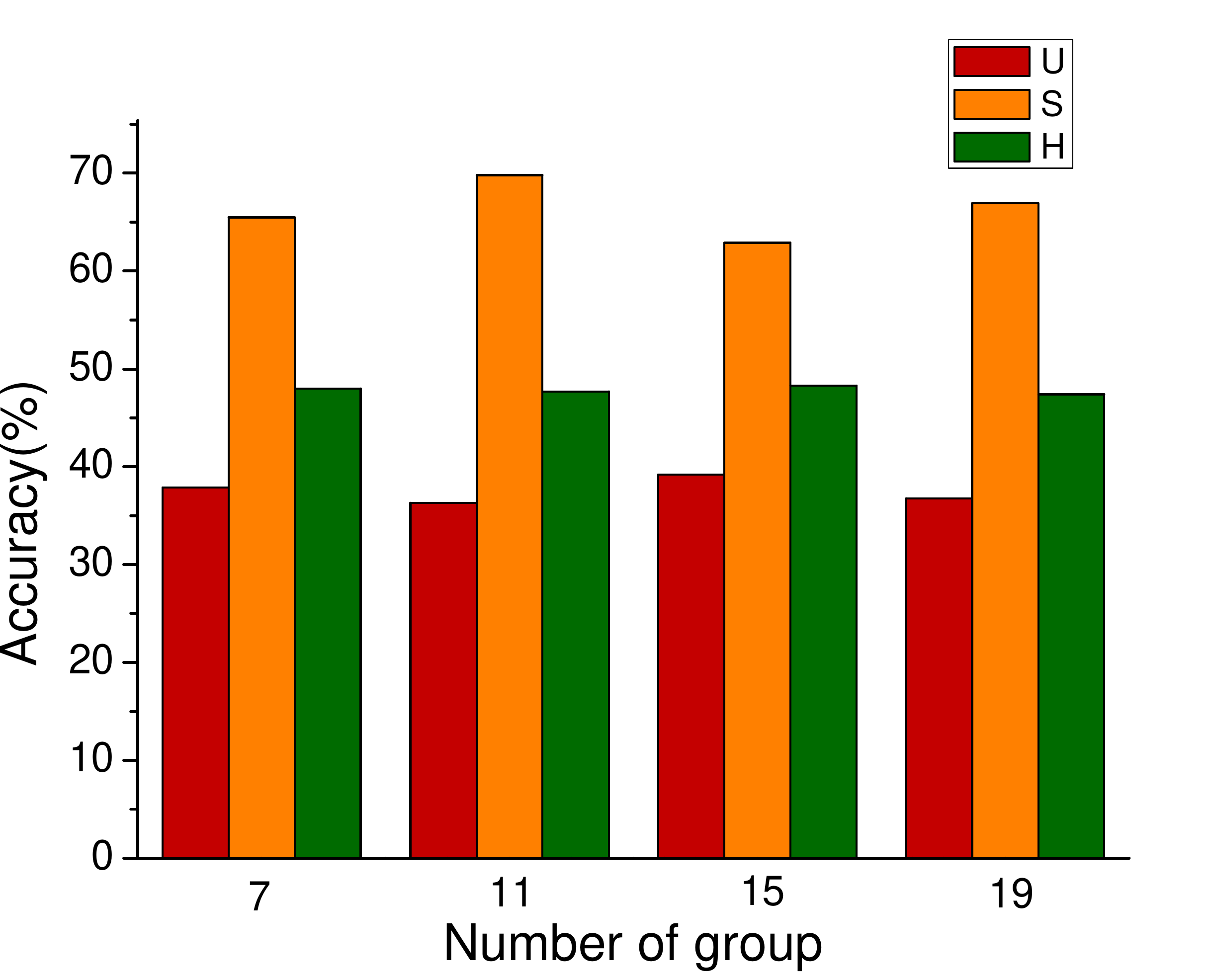}	
		\end{minipage}%
	}%
	\subfigure[CUB]{
		\begin{minipage}[t]{0.25\textwidth}
			\centering
			\includegraphics[width=1\columnwidth]{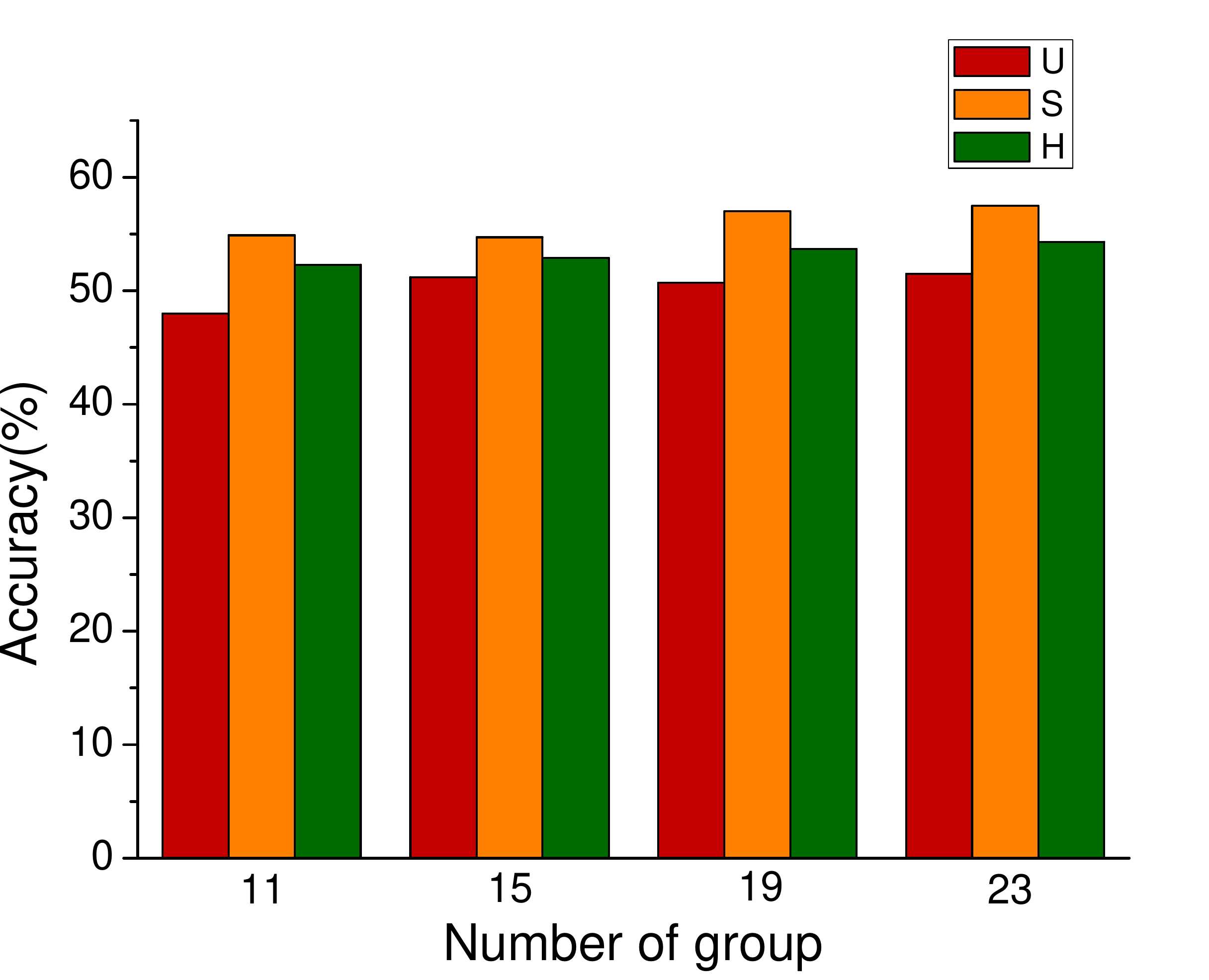}	
		\end{minipage}%
	}%
	\subfigure[SUN]{
		\begin{minipage}[t]{0.25\textwidth}
			\centering
			\includegraphics[width=1\columnwidth]{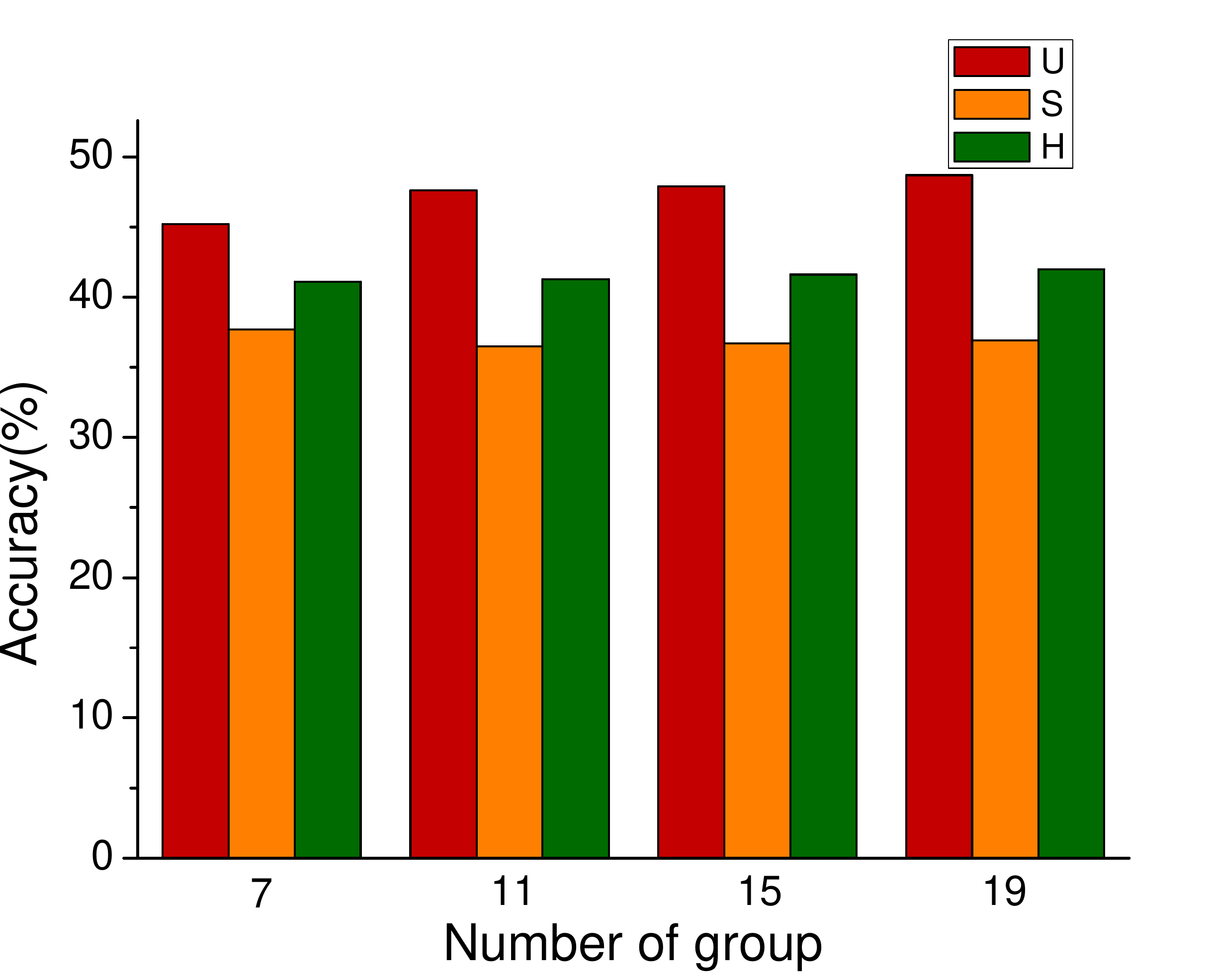}	
		\end{minipage}%
	}%
	\vspace{-2ex}
	\centering
	\caption{The influence of different grouping numbers on GZSL results in attribute clustering.}
	\vspace{-4ex}
	\label{fig3}
\end{figure*}

\section{Experiments}

\subsection{Datasets Introduction and Setting}
We evaluated our method on four data sets, \textit{i.e.}, AWA1 \cite{lampert2009learning} containing 50 animal category attribute descriptions; CUB \cite{welinder2010caltech} is a fine-grained data set about birds; SUN \cite{patterson2014sun} is a fine-grained data set about visual scenes; Attribute Pascal and Yahoo are abbreviated as aPY \cite{farhadi2009describing}. Table \ref{table3} shows the number division of seen and unseen classes, attribute dimensions and number of samples of the four data sets. All data sets extract visual features of 2048 dimensions through ResNet-101 \cite{he2016deep} without finetuning. The semantic attributes of the four data sets adopt the original semantic attributes.


We implement the proposed method on PyToch.  We use a layer of network with Sigmod activation function for Self-supervision classifier. After training the generation network, for the visual features generated by unseen classes, on the basis of using a single category attribute, we use the incomplete semantic attributes of classes to generate some visual features in proportion. At the same time, the visual features generated by our generation network are more diversified. In order to prevent over fitting of the generation network, we add L2 regularization to the optimizer of the generation network.

\begin{table}[!t]
	\centering	
	\begin{tabular}{lllll}
		\hline
		Datasets & Seen & Unseen & Attribute & Samples \\ \hline
		AWA      & 40   & 10     & 85        & 30,475\\
		aPY      & 20   & 12     & 64        & 15,339\\
		CUB      & 150  & 50     & 312       & 11,788\\
		SUN      & 645  & 72     & 102       & 14,340\\ \hline
	\end{tabular}
	\vspace{-1.5ex}
	\caption{Division of seen and unseen classes, attribute dimensions and number of samples of four data sets.}
	\vspace{-3ex}
	\label{table3}
\end{table}

\subsection{Evaluation Protocols}
We follow the evaluation method proposed by harmonic average \cite{xian2018zero}. For GZSL, we calculate the average accuracy $S$ and $U$ of seen and unseen classes respectively. The performance of GZSL is evaluated by their harmonic average: $H=2\times S\times U / \left (S+U  \right )$.

\begin{figure}[ht]
	\centering
	\subfigure[RFF-GZSL]{
		\includegraphics[width=0.22\textwidth]{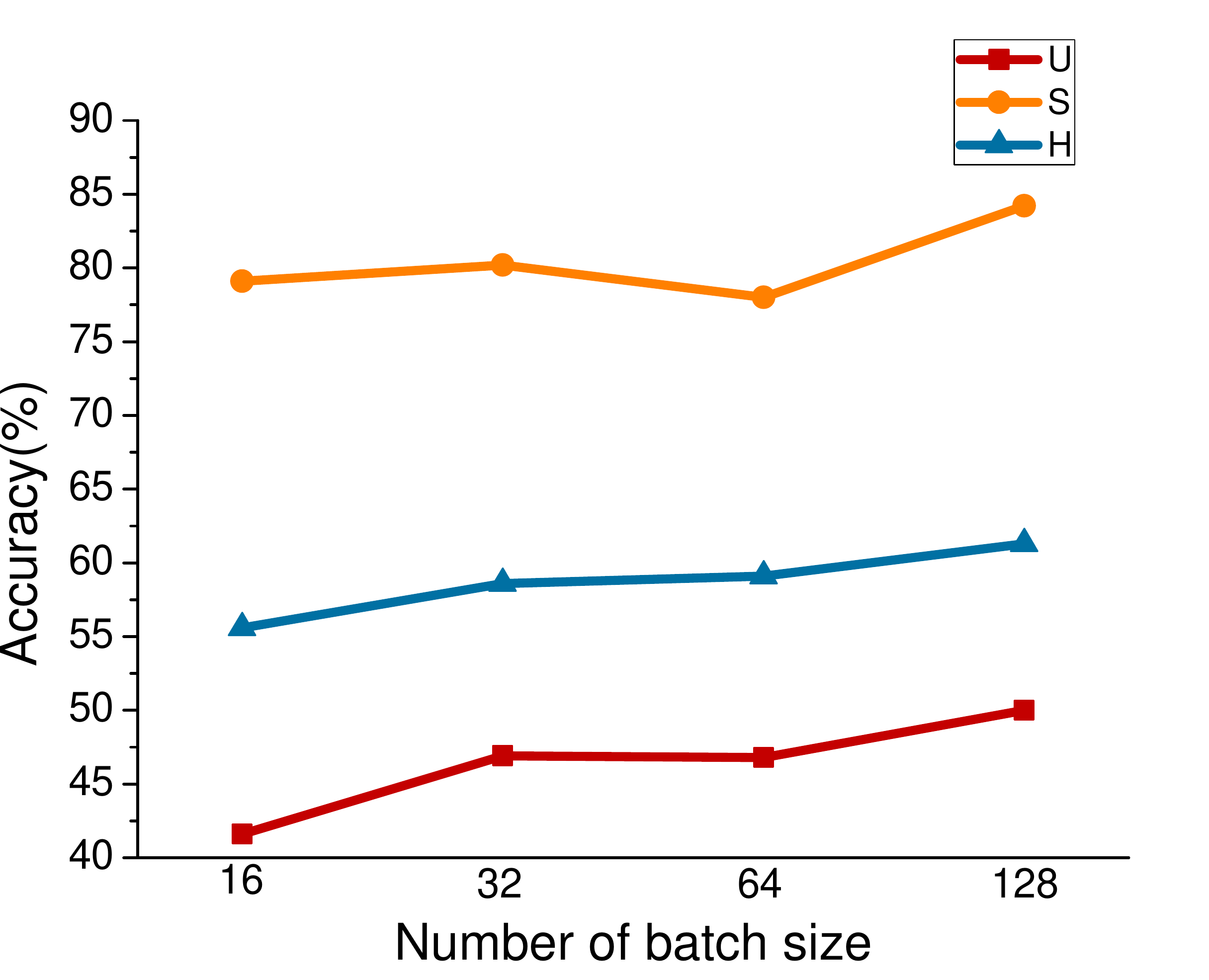}}%
	\subfigure[RFF-GZSL+SDFA$^{2}$]{
		\includegraphics[width=0.22\textwidth]{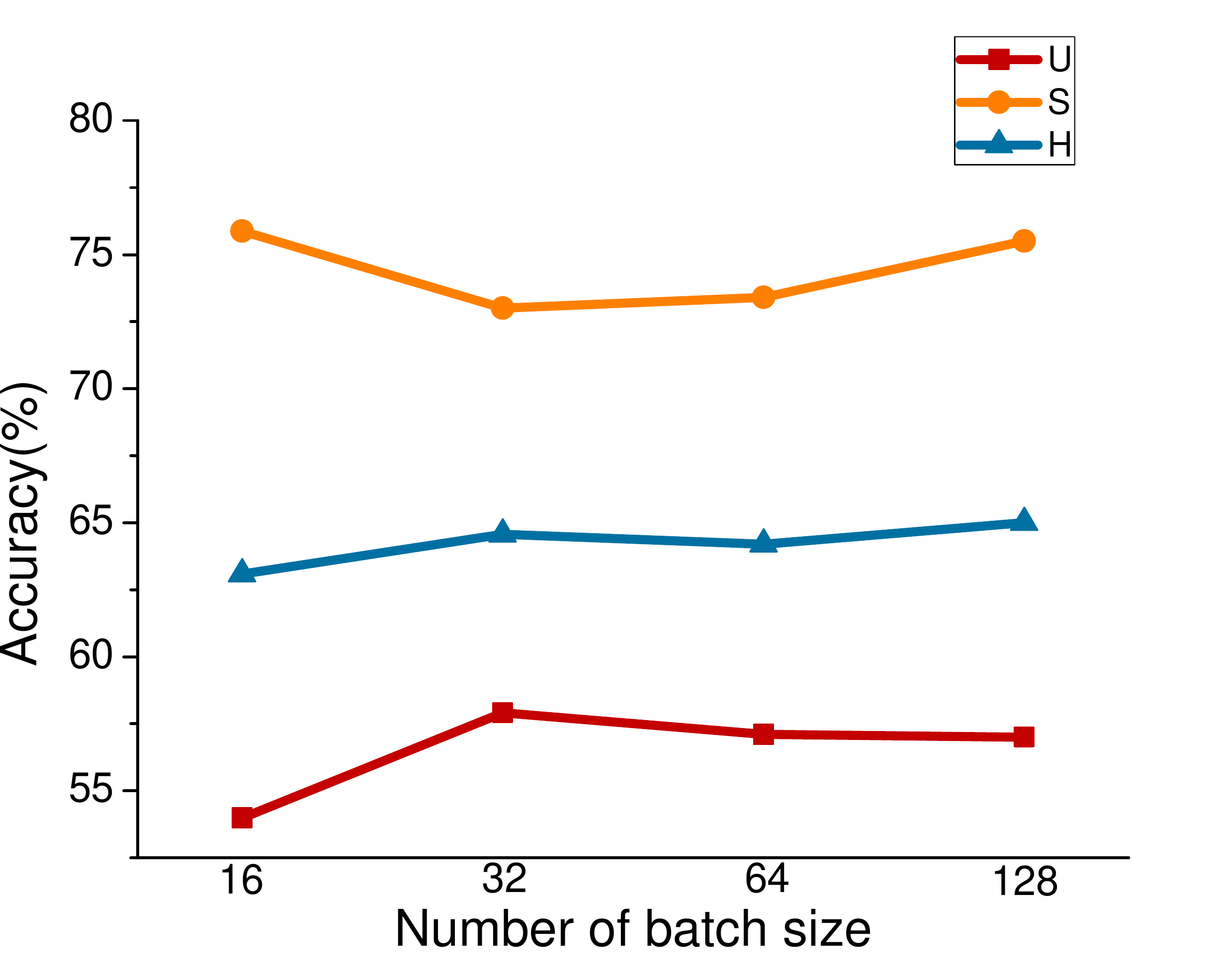}}
		\vspace{-2ex}
	\caption{Effects of different batch sizes on experimental results.}
	\vspace{-3ex}
	\label{fig6}
\end{figure}

\subsection{Improvement against Existing Methods}
In order to verify that our incomplete generation method is a general method, we add our generation method based on the models of f-WGAN, f-VAEGAN-D2, RFF-GZSL and CE-GZSL.

For f-CLSWGAN, a single attribute is used to generate features and calculate generative adversarial loss and classification loss. On this basis, after using our method to generate diversity features, we need to calculate not only the corresponding generative adversarial loss, but also the loss of the generated diversity features on the pre trained classifier.

For f-VAEGAN-2D, the original model is divided into VAE and WGAN. SDFA$^2$ only modifies WGAN module. On the basis of the original single attribute generation feature, which produces the generative adversarial loss, increases the generative adversarial loss caused by using the diversity generation feature.

For RFF-GZSL, the additional features generated by our method also need to go through mapping function to transform the diversity visual features to the redundancy free features, and then calculate the corresponding loss. Similarly, the original model adopts the classification loss in f-CLSWGAN, and the diversity features we generate also need to calculate the classification loss.

For CE-GZSL, the additional generated diversity visual features need to go through the embedding function proposed by the original model, and the corresponding loss is calculated through two modules: instance-level contrast embedding and class-level contrast embedding.

\subsection{Parameter Setting}

Due to the different experimental equipment and parameter settings, which most important thing is that we only have one GTX 1080ti GPU, it is difficult to restore the test results given by some generation models. However, we just want to verify that our method is a general method for generating class model, so we use the method of controlling variables.

For RFF-GZSL, we change the batch size. We set batch size to 128 on all four data sets. At the same time, for the fine-grained data sets SUN and CUB, we use the original attributes. Other parameters refer to the values given in the paper. For CE-GZSL, we hope to use the batch size provided in the paper, but the model is too complex and limited by equipment. We also set batch size to 128, which will have a certain impact on the comparative learning module in CE-GZSL. Other parameter settings follow the values given in the paper. In order to prove the influence of different batch sizes on the experimental results, we set different batch sizes on RFF-GZSL. The results are shown in Figure \ref{fig6}, we can clearly see that $H$ is increasing with the increase of batch size.

On the basis of following the parameters given by these three models, we obtain the operation results on our equipment. Then, we fix all the parameters of the model without changing, and add our proposed method. The experiment in this section is not to compare with the state-of-the-art methods, but to verify that the general method proposed by us is universal in improving the effect of the generated model. The experimental results are shown in table \ref{table1}.

\subsection{Comparison and Evaluation}
It can be seen that our method has significantly improved the effect of the three generation methods. Especially on the AWA1 and aPY data sets, the improvement effect is obvious. See the f-CLSWGAN with SDFA$^{2}$, the AWA1 is increased by 5.6 percentage points, and the aPY is increased by 4.5 percentage points. On f-AVEGAN-D2, after using SDFA$^{2}$, the H of AWA1 is 64.7\% and aPY is 46.8\%. After using SDFA$^{2}$, CE-GZSL and RFF-GZSL increased by 1.5\% and 3.5\% respectively for AWA1, and increased by 17\% and 8.3\% respectively for aPY. It can be seen that the features of generating diversity contribute to the improvement of performance.

For the fine-grained data sets SUN and CUB, because the gap between classes is subtle, the gap between categories becomes smaller after masking a feature to 0, so the effect improvement is not as significant as AWA1 and aPY. However, the improvement in f-CLSWGAN and f-AVEGAN is still significant, reaching 42.0\% and 42.4\% in SUN, 54.2\% and 54.2\% in CUB respectively. There are also some improvements in CE-GZSL and RFF-GZSL.
\begin{table*}[t]
	\centering
	\begin{tabular}{l|ccc|ccc|ccc|ccc}
		& \multicolumn{3}{c|}{\textbf{AWA1}} & \multicolumn{3}{c|}{\textbf{aPY}} & \multicolumn{3}{c|}{\textbf{SUN}} & \multicolumn{3}{c}{\textbf{CUB}} \\
		& U          & S         & H         & U         & S         & H         & U         & S         & H         & U         & S         & H        \\ \hline
		f-CSLWGAN+$L_{div-gan}$            & 58.9       & 71.3      & 64.5      & 37.0      & 64.7      & 47.1      & 47.2      & 35.4      & 40.5      & 51.7        & 56.7        & 54.1       \\
		f-CLSWGAN+$L_{div-gan}$+$L_{self}$ & 59.1       & 72.8      & 65.2      & 38.0      & 62.8      & 47.4      & 48.7      & 36.9      & 42.0      & 51.5        & 57.5        & 54.3       \\ \hline
	\end{tabular}
	\vspace{-1ex}
	\caption{The table shows the ablation experiments of diversity loss $L_{div-gan}$ and self-supervision loss $L_{self}$ on f-CLSWGAN.}
	\vspace{-1ex}
	\label{table2}
\end{table*}

\subsection{Number of Attribute Groups}
In the process of attribute grouping, the selection of the number of groups, in other words, the selection of the number of cluster centers is a problem to be considered. The choice of different number of cluster centers will have different effects on the final results. In order to test the influence of attribute grouping on diversity generation, we set up different number of clustering centers in the process of attribute clustering. Figure \ref{fig3} shows the results of different data sets based on f-CLSWGAN with SDFA$^{2}$ after setting different clustering centers. We can clearly see that the number of different attribute groups has little effect on the performance of the model.


\begin{figure*}[ht]
	\subfigure[Single attribute generation method.]{
		\includegraphics[width=0.28\textwidth]{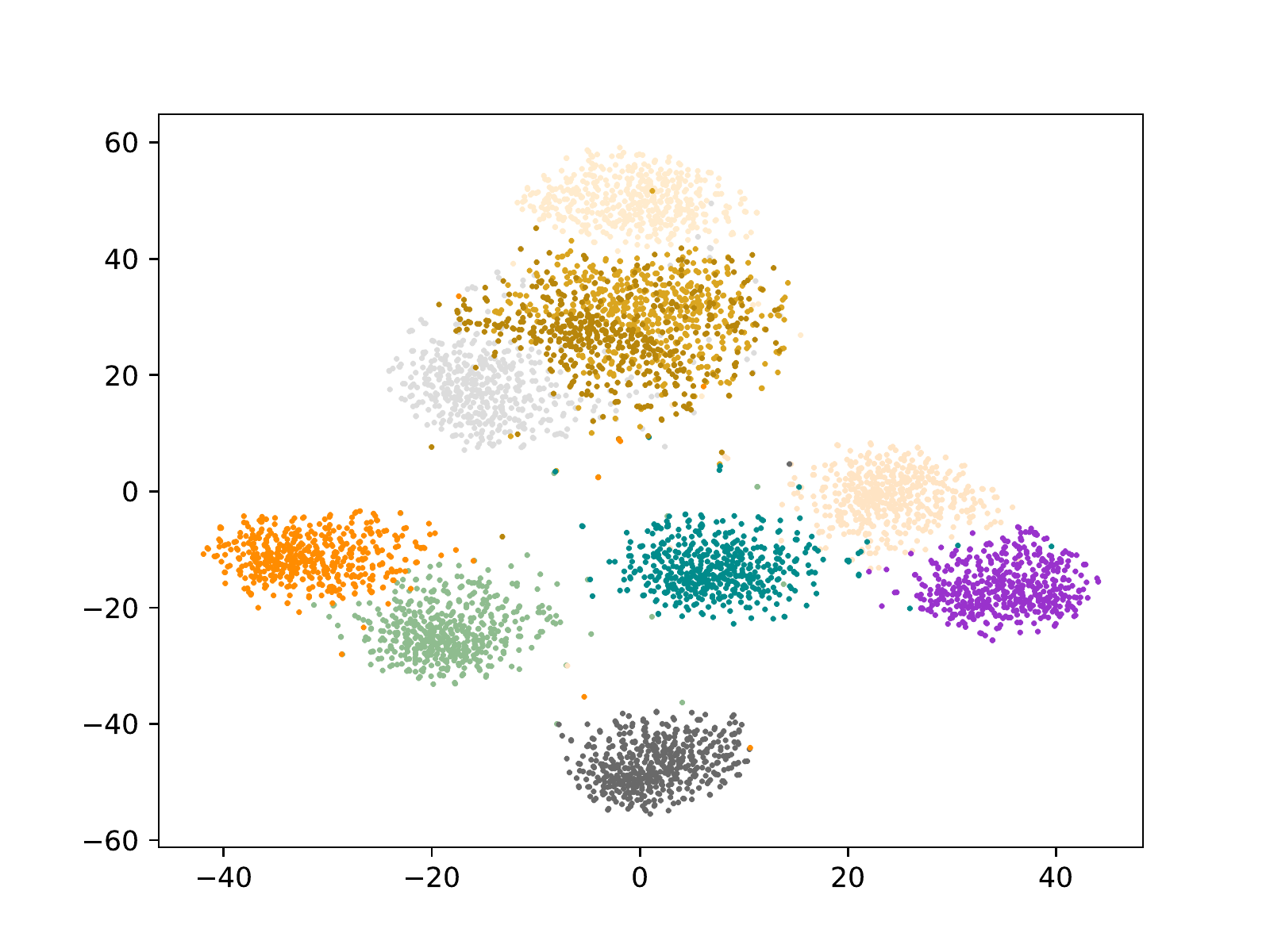}
	} \qquad
	\subfigure[Real	features distribution.]{
		\includegraphics[width=0.28\textwidth]{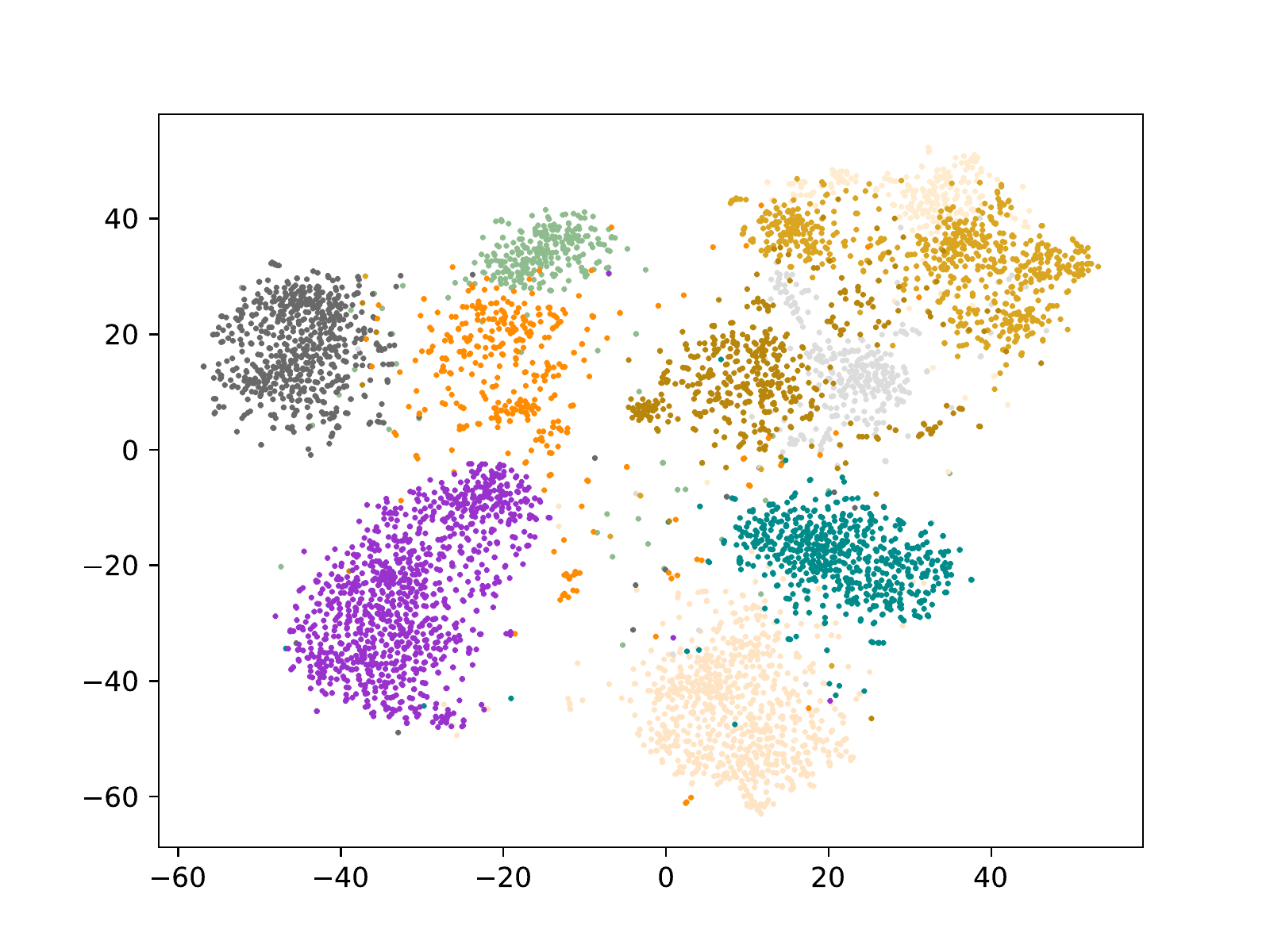}	
	}\qquad
	\subfigure[SDFA$^{2}$.]{
		\includegraphics[width=0.28\textwidth]{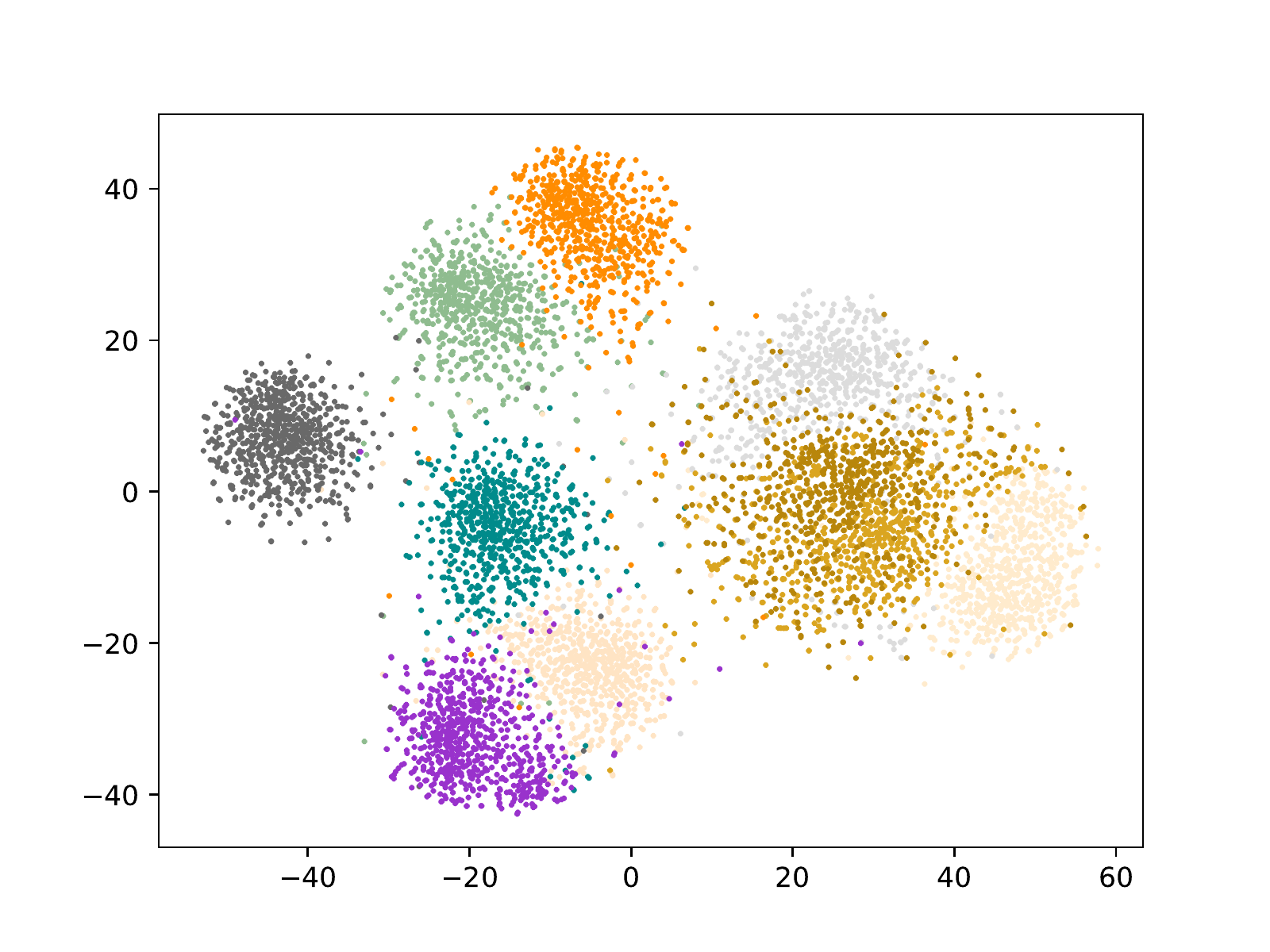}	
	}%
	\centering
	\vspace{-2ex}
	\caption{t-SNE graph of visual features generated by single attribute and diversity attribute generation method.}
	\vspace{-3ex}
	\label{fig5}
\end{figure*}

\subsection{Diversity Generation}
The main purpose of our proposed SDFA$^{2}$ is to generate visual features through diversity attributes. Different from the method of generating visual features using a single attribute, which only increases diversity through Gaussian noise. Figure \ref{fig5} (a) shows the results of the visual features generated by the traditional single attribute generation method under t-SNE. It can be seen that the generated visual features show an obvious Gaussian distribution.
Figure \ref{fig5} (b) shows the features distribution of samples of real unseen classes. Figure \ref{fig5} (c) is the result of visual features generated by the generation network trained using SDFA$^{2}$ method under t-SNE. Obviously, it is more in line with the real distribution. So, the generated samples can better replace the real samples for classifier training.

\subsection{Ablation Experiment}
In order to verify the impact of incomplete generation method and self supervised classification loss on the performance of the model, we designed ablation experiments. We first increase only diversity loss, then add diversity loss and self-supervision loss at the same time. The experimental results are shown in Table \ref{table2}. It can be seen that incomplete loss and self-monitoring loss significantly improve the performance of the model.
\begin{figure}[ht]
	\centering
	\vspace{-2ex}
	\subfigure[AWA1]{
		\includegraphics[width=0.21\textwidth]{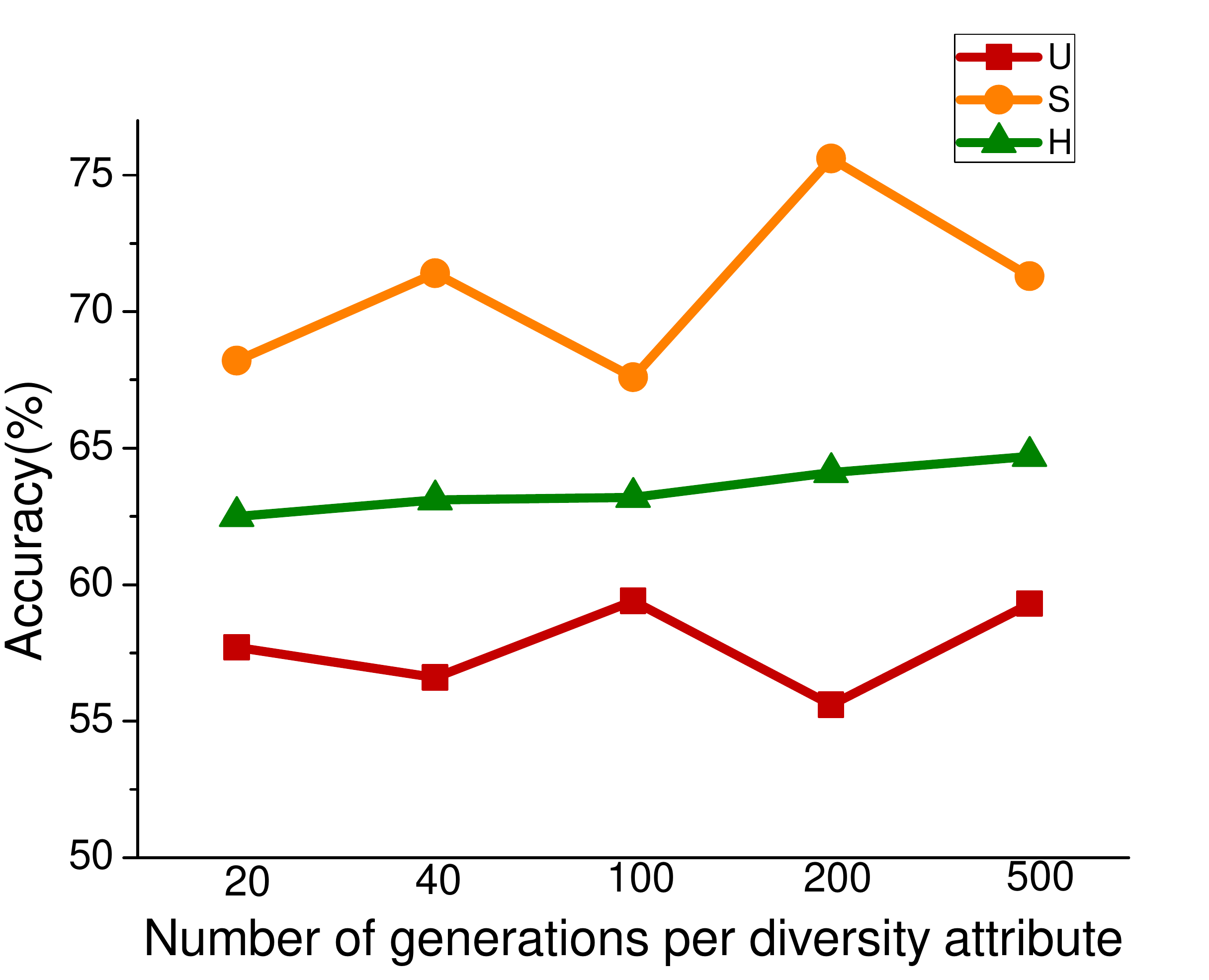}}\quad
	\subfigure[aPY]{
		\includegraphics[width=0.21\textwidth]{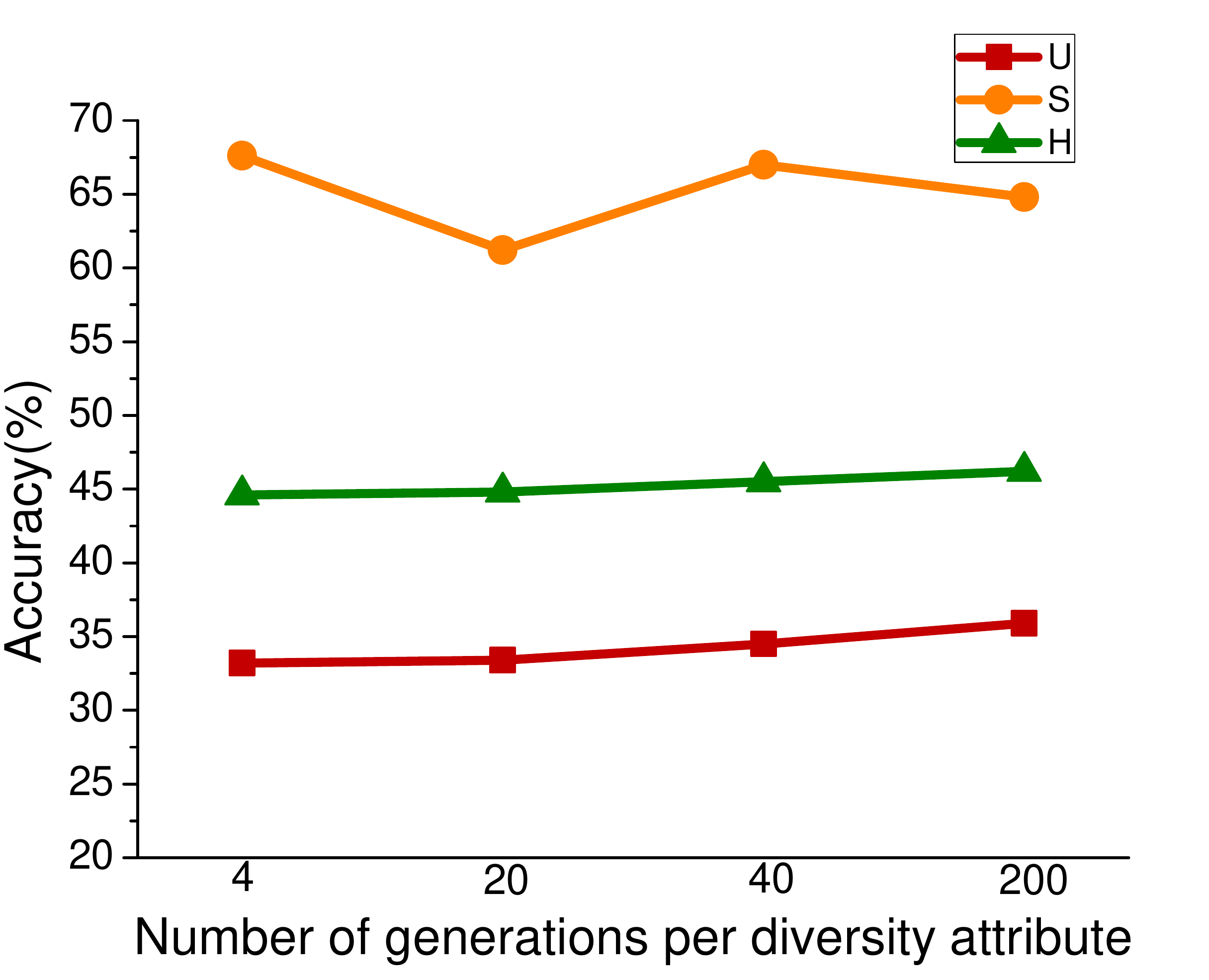}} \\
		\vspace{-2ex}
	\subfigure[CUB]{		
		\includegraphics[width=0.21\textwidth]{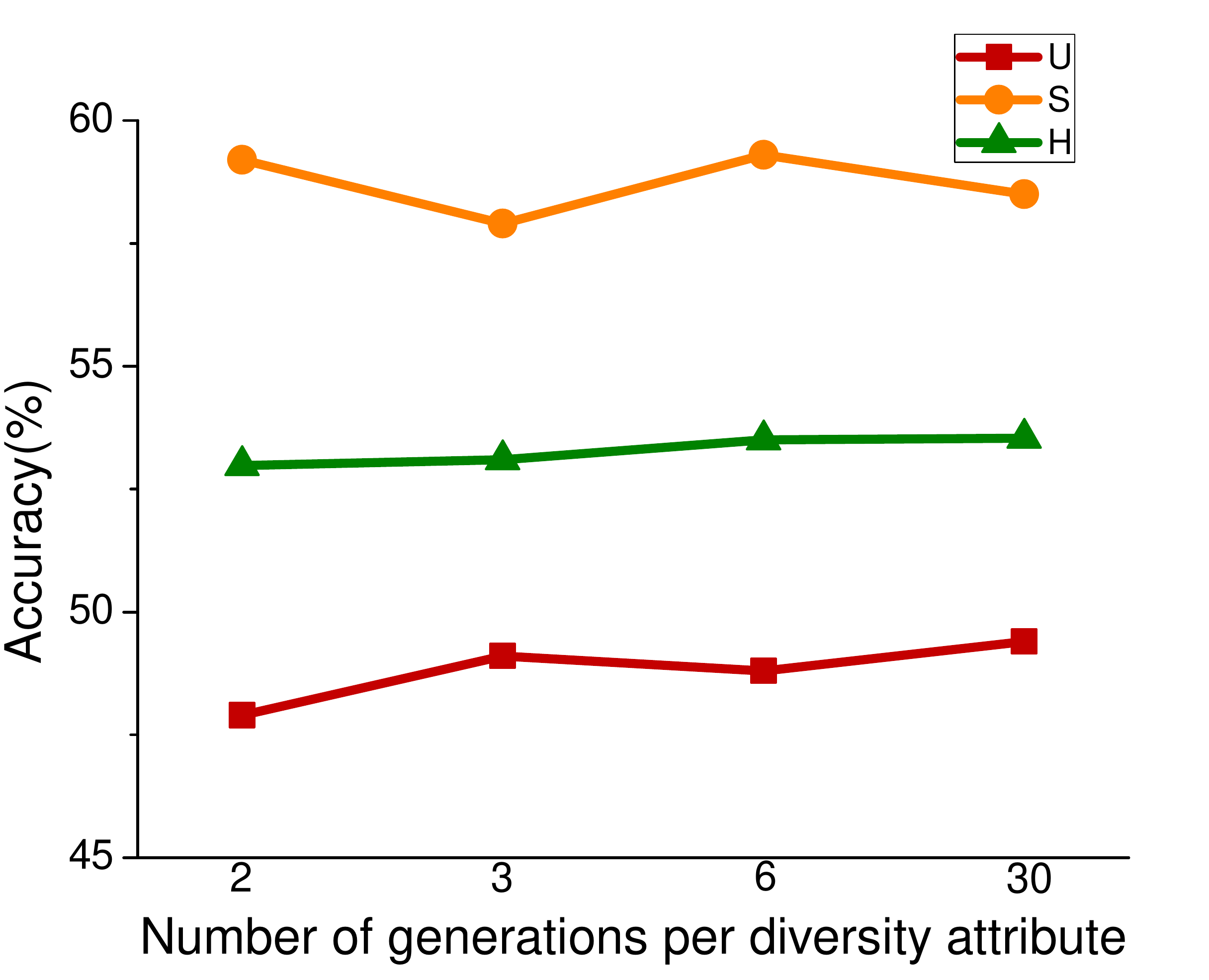}}\quad
	\subfigure[SUN]{
		\includegraphics[width=0.21\textwidth]{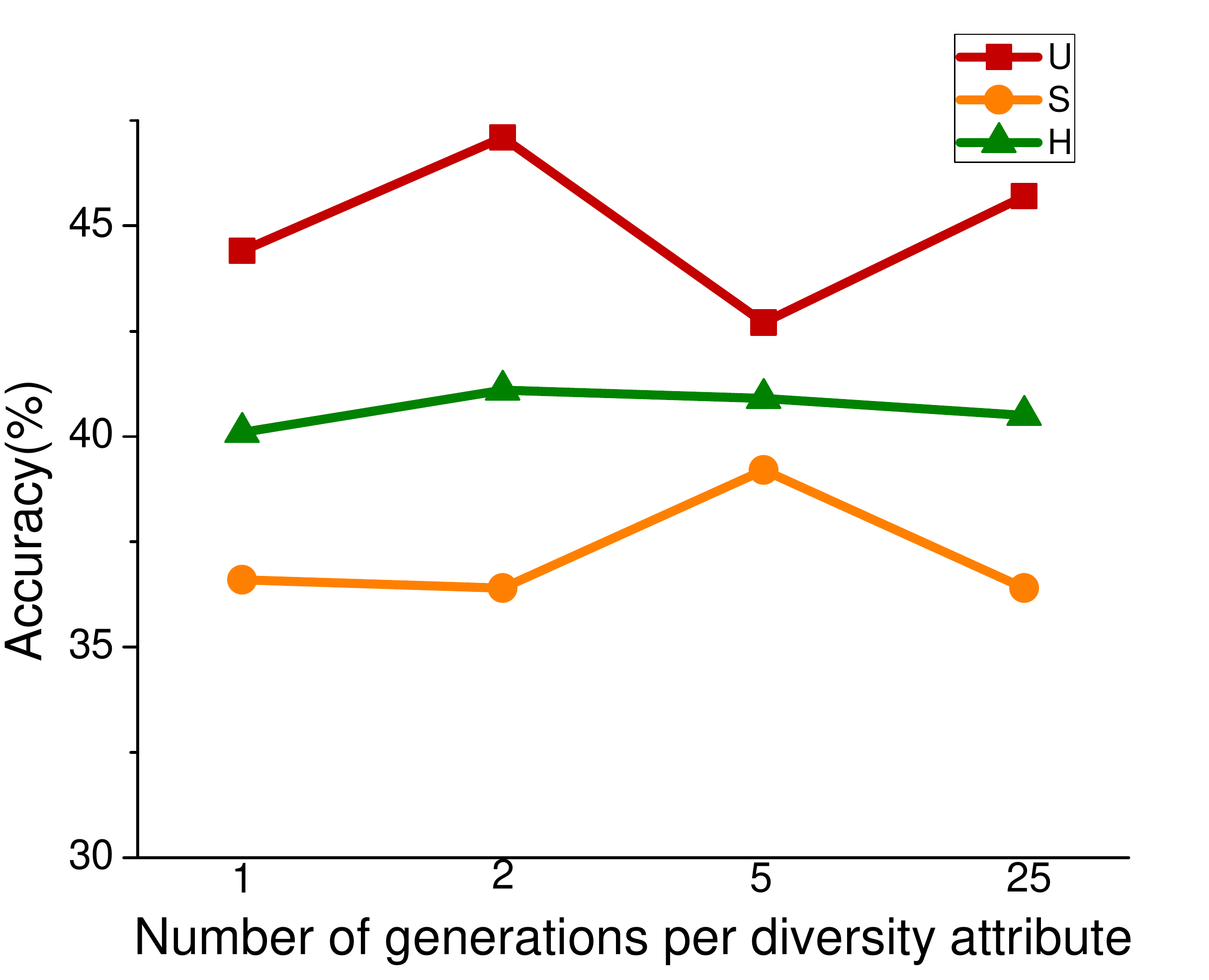}}%
		\vspace{-3ex}
	\caption{The impact of generating different numbers of diverse features on the results.}
	\vspace{-3ex}
	\label{fig4}
\end{figure}
\subsection{Numbers of Synthesized Features}
The number of diversity generation is an important parameter in our method. In order to explore the impact of different generation quantities on the performance of the model, we designed this experiment. We fixed the number of visual features generated by a single attribute for each class, AWA1 with 2000, aPY with 2000, CUB with 300 and SUN with 50. Then, according to different proportions, the diversity attribute is used to generate additional visual features. Figure \ref{fig4} shows the results of four data sets at different scales. It is obvious that there are significant differences in the results under different proportions. This is because in the real situation, there is a certain relationship between the number of pictures with complete attributes and the number of pictures with incomplete attributes, so the number of generated pictures with different proportions will inevitably have a certain impact on the results.

\section{Conclusion}
In this work, we propose an attribute enhancement method to generate diversity features, which is used to drive the generated feature distribution to approach the real feature distribution. We solve the problem that the attribute features used in the existing generative GZSL methods are too single. At the same time, based on diversity feature generation, a self-supervision loss is proposed to enhance visual feature generation. Experiments show that our SDFA$^{2}$  has good generality on different generative GZSL models. Considering that attribute mask is not the only way to generate diversity attributes, we will do more processing on attributes in external work in order to generate visual features more in line with the real distribution.

\section{Acknowledgments}
This work was supported in part by the National Natural Science Foundation of China (NSFC) under Grants No. 61872187, No. 62077023 and No. 62072246, in part by the Natural Science Foundation of Jiangsu Province under Grant No. BK20201306, and in part by the ``111 Program'' under Grant No. B13022.

\bibliography{reference}

\end{document}